\documentclass[11pt]{article}
\title{A Spectrum of Applications of Automated Reasoning\thanks{
This work was supported by the Mathematical, Information, and Computational Sciences Division subprogram of the Office of Advanced Scientific Computing Research, U.S. Department of Energy, under Contract W-31-109-Eng-38.}}

\author{Larry Wos\\
        Mathematics and Computer Science Division\\
           Argonne National Laboratory\\
           Argonne, IL  60439-4801\\
            e-mail: wos@mcs.anl.gov}
\date{January 2002}

\begin{document}
\maketitle

\begin{abstract}

The likelihood of an automated reasoning program being of substantial assistance for a wide spectrum of applications rests with the nature of the options and parameters it offers on which to base needed strategies and methodologies.
This article focuses on such a spectrum, featuring W. McCune's program OTTER, discussing widely varied successes in answering open questions, and touching on some of the strategies and methodologies that played a key role.
The applications include finding a first proof, discovering single axioms, locating improved axiom systems, and simplifying existing proofs.
The last application is directly pertinent to the recently found (by R. Thiele) Hilbert's twenty-fourth problem---which is extremely amenable to attack with the appropriate automated reasoning program---a problem concerned with proof simplification.
The methodologies include those for seeking shorter proofs and for finding proofs that avoid unwanted lemmas or classes of term, a specific option for seeking proofs with smaller equational or formula complexity, and a different option to address the variable richness of a proof.
The type of proof one obtains with the use of OTTER is Hilbert-style axiomatic, including details that permit one sometimes to gain new insights.
We include questions still open and challenges that merit consideration.
\end{abstract}
\section{Background and Perspective} %
 
In this article, we make a strong case for the use in diverse applications of an automated reasoning program by mathematicians and logicians.
The basis consists of brief discussions (in Section 3) of successes in answering open questions from unrelated fields and of finding missing proofs of various types.
(Numerous missing proofs and input files will be offered in the forthcoming book entitled {\em Automated Reasoning and the Discovery of Missing and Elegant Proofs} by L. Wos and B. Fitelson.)
In this section, we set the stage for presenting what can be accomplished with McCune's automated reasoning program.
(For a detailed treatment of automated reasoning, see [Wos1999].)
Especially for those not familiar with this program or, more generally, with the field, we provide in this section a somewhat detailed example of how a first and significant proof was discovered.
 
Perhaps the most difficult task for a mathematician or logician, and yet clearly intriguing and pleasurable, is that of proof finding.
Precisely how such a researcher completes proofs remains a mystery.
What is clear is that various proofs sometimes are missing for many decades, as was the case with the proof that every Robbins algebra is a Boolean algebra (proved by W. McCune with his automated reasoning program EQP [McCune1997]).
 
Proofs take many forms, including those by induction, those relying on some very powerful result such as Zorn's lemma or the well-ordering principle, those that are purely first-order and axiomatic in the style of Hilbert, and those proofs by contradiction.
Our preference is for Hilbert-style axiomatic proofs that are purely first-order and that complete by detecting a contradiction.
In our view, compared with other types of proof, such an axiomatic proof is more likely to provide new insights and is far more instructive in general.
Indeed, one can learn from such a proof.
 
We preferred such proofs even as early as the mid-1950s when in the mathematics department at the University of Chicago, and we still do.
Therefore, our fascination with automated reasoning and the proofs discovered by McCune's program OTTER comes as no surprise.
This article cites (in Section 3) such proofs with little detail, proofs that answer diverse open questions taken from a variety of fields of mathematics and logic.
 
The type of attack OTTER applies in general differs sharply from that of the typical unaided researcher; no attempt is made to emulate some master of some field.
Instead, when a deep question or hard problem is under consideration, the program ordinarily accrues a vast amount of new conclusions with the objective of finding among them a contradiction.
On the other hand, the program does not undertake a study on its own.
Rather, especially in our research, a form of advice is usually given and (one hopes) wise choices are made for the options used and effective choices are made for the values assigned to the parameters.
The following example illustrates to a small extent how we attack a problem, how we search for a missing proof.
 
The theorem of concern, actually the proof, focuses on two-valued sentential (or propositional) calculus.
Whereas the Robbins problem featured three equations and asked whether they provided an axiomatization of Boolean algebra, the focus here is on a single formula (not relying on equality) and the assertion that it {\em does} provide an axiomatization for propositional calculus.
Specifically, in the mid-1930s, J. {\L}ukasiewicz offered without proof the following 23-letter formula, where the function $i$ denotes implication and the function $n$ negation, and noted that it sufficed for the study  of the cited area of logic [{\L}ukasiewicz1970].
\begin{verbatim}
i(i(i(x,y),i(i(i(n(z),n(u)),v),z)),i(w,i(i(z,x),i(u,x))))
\end{verbatim}
Our consultations with colleagues strongly suggested that a proof of this fact had never been published.
In other words, sharing with the Robbins problem, a proof was missing.
Our goal was to find such a proof, a proof that showed the {\L}ukasiewicz 23-letter formula to be a single axiom for propositional calculus.
Also similar to the Robbins problem, one was evidently free to choose the target for the desired proof.
 
In contrast to a conjecture, we were certain that a proof had existed; after all, {\L}ukasiewicz was a master.
The question remained regarding what target he had in mind.
We chose as target his three-axiom system [{\L}ukasiewicz1970], the following.
\begin{verbatim}
i(i(x,y),i(i(y,z),i(x,z)))
i(i(n(x),x),x)
i(x,i(n(x),y))
\end{verbatim}
As noted earlier, we usually give OTTER suggestions for an attack.
One way for the researcher to do this is to include an appropriate list equations or formulas that the researcher considers attractive because of their shape.
The variables of such included items, called {\em resonators} [Wos1995], are treated as indistinguishable, thus making their functional shape the key.
To each resonator, one assigns a value to reflect the conjectured importance of the pattern: the lower the value, the higher the priority given to any deduced conclusion that matches the corresponding resonator.
To direct its reasoning, the program can be instructed to choose from among its database of conclusions that which has the highest priority.
Perhaps influenced by our choice of target (the {\L}ukasiewicz three-axiom system), we included sixty-eight resonators, each corresponding to a thesis (theorem) that {\L}ukasiewicz had included in his publications [{\L}ukasiewicz1970].
We assigned to each a very small value to give any deduced conclusion matching one of the sixty-eight resonators a priority (higher than any other conclusion) for initiating the application of an inference rule.
 
The second important aspect of our methodology was that of temporary lemma adjunction [Wos2001a].
The lemmas to be adjoined, if proved, were from among the sixty-eight theses.
Those that were proved in one run were adjoined in the initial set of support for the next run.
The style of the methodology was interactive.
 
The third aspect of the methodology concerned the inclusion in later runs of resonators corresponding to proof steps of lemmas proved in earlier runs.
In addition to proved lemmas among the cited sixty-eight, proof steps of any of the target axioms were included (in the initial set of support) if proved.
As it turned out, the third of the three {\L}ukasiewicz axioms was proved in an early run and the second proved in the next run.
 
The final aspect---and one that we conjecture enabled the program to succeed---was most counterintuitive, if one examines the literature.
Specifically, for all runs we instructed the program to avoid retention of any conclusion that contained a double-negation term, a term of the form $n(n(t))$ for any term $t$.
That decision was motivated by three factors.
First, for many years such prohibition had proved to be most effective in proof finding with OTTER.
Second, we were curious about the possible existence of a double-negation-free proof of this marvelous theorem.
Third, we had come to believe that the density of proofs within the space of double-negation-free conclusions was far greater than in the entire space of conclusions.
 
Of course, consistent with our preference for remaining strictly within the theory under study, we excluded any mention of equality and confined the inference-rule mechanism to condensed detachment.
A glance at the work of various masters shows that, in cases of the type under discussion, equality is sometimes brought into the picture.
Our goal was to complete a proof relying solely on condensed detachment, conjecturing that such a proof in general provides more insight and is often easier to follow.
 
In three runs, OTTER produced a proof of the three-axiom system of {\L}ukasiewicz.
Because that proof relied upon various lemmas adjoined during the process, it was not quite what we were after.
Indeed, the proof produced in the third run nicely established that the goal was reachable.
Therefore, in the fourth run, all temporarily adjoined lemmas were removed.
Two sets of resonators were included, one corresponding to the key proof found in the third experiment and one corresponding to proof steps of lemmas from among the sixty-eight not proved in earlier runs.
 
In contrast to the third experiment's heavy use of CPU time, the fourth experiment quickly completed, yielding a 200-step proof [Wos2001a].
Its length and its nature (free of double negation) almost certainly guarantee that the original and unpublished and unavailable {\L}ukasiewicz proof was in no way similar to that produced by OTTER---we shall never be able to make that interesting comparison.
Few if any would enjoy a close examination of a 200-step proof.
Besides, pertinent to the Hilbert twenty-fourth problem, a vigorous attempt was in order to find a far, far shorter proof.
We therefore undertook the needed investigation.
More than one year of not continuous study witnessed progress---a 50-step proof was discovered [Wos2002].
We offer as a challenge the finding (if such exists) of a proof of length strictly less than fifty.
We place no constraint on the target; in particular, the {\L}ukasiewicz three-axiom system need not be the choice.
\section{Solvable Problem Classes} %
 
Two factors explain the content of this section, namely, the sampling of some of the types of problem that are amenable to attack with OTTER.
First, we wish to inform the various researchers about what can be accomplished, the diversity that is accessible.
Second, we continue to seek open questions, hard problems, and proofs that merit refinement.
Therefore, a discussion of the types of problem that can be attacked is in order; we seek problems in one or more of the classes we discuss in this section.
To further clarify the type of problem most amenable to study with OTTER, we touch on various methodologies this program supports that have proved quite powerful.
Almost always, an assignment is completed by finding a Hilbert-style axiomatic proof by contradiction.
The researcher includes as part of the input a statement or statements that correspond to assuming the theorem false or the assignment uncompletable.
 
To many, the most attractive class of problem concerns finding a {\em first proof}, which may be in the context of settling a conjecture or of producing a proof for a result announced without proof.
The approach we take in such cases generally focuses on searching where no researcher has gone before.
Indeed, at least for conjectures, we do not expect to improve upon the work of an expert's exploration of a given terrain.
Therefore, we often make counterintuitive moves such as avoiding double negation or avoiding some previously-thought-to-be crucial lemma.
Such avoidance is effected by the use of demodulation, rewriting the unwanted to junk to be purged, or by means of weighting, assigning the unwanted a complexity that exceeds the assigned value (by the user) for the complexity of newly retained information.
 
With or without such counterintuitive moves, the program still provides the basis for actions that the unaided researcher might find impractical to take.
Indeed, one can instruct OTTER to retain extremely complex conclusions (measured in symbol count) by assigning the max\_weight a correspondingly high value.
Further, one can instruct the program to focus on such complex conclusions by simply choosing a breadth-first search, set(sos\_queue).
Most unlikely is the case in which an unaided researcher would find such a search practical.
One can modify the breadth-first search by mixing it with a complexity-driven search by relying on McCune's ratio strategy, which blends the two direction strategies according to the value assigned to pick\_given\_ratio.
Also, as in part discussed in Section 1, the researcher can advise the program about which equational patterns or which formula patterns are attractive by using the resonance strategy.
Many, many additional actions can be taken to direct the program's reasoning or to restrict it in search of a first proof or in the attempt to settle a conjecture.
Here we have merely provided a small taste.
 
Of a related nature is the seeking of {\em single axioms} for some area or the seeking of a {\em preferable axiom system or basis}.
In such cases, one turns to the same means as cited for seeking a first proof.
One can, however, do as we do when wishing to consider many combinations of parameter values and option settings.
Specifically, we use super-loop, a program that considers all of the combinations dictated by a user-supplied addendum to an input file, and we use otter-loop, a program that automatically runs a sequence of experiments that differ by, for example, blocking the use of one step of a proof after another.
 
Problems of axiom dependence are often easily solved with OTTER.
Sometimes one can off-and-on study an area in terms of an axiom system and be unaware that dependencies exist among its members.
One aspect of mathematics and logic focuses on learning about such dependencies, such as the dependency of the axioms of right identity and of right inverse in group theory.
In the mid-1990s, the logician R. Epstein [Epstein1995] offered an open question on axiom dependence for a six-axiom system for propositional logic, a question quickly answered by OTTER.
The approach to the study of possible axiom dependence is straightforward.
One places all but one of the members in the initial set of support, places the negation of the remaining member in the passive list, and seeks a proof.
Each member is successively treated in this manner.
Of course, semidecidability comes into play; indeed, if no proof is found, one cannot be certain that a corresponding dependency does not exist.
This situation does not differ materially from that in which a colleague is asked for a proof and does not deliver such.
When a proof is not forthcoming and doubt begins to grow, one can turn to some model generation program.
 
We now turn to open questions of a different type, namely, those concerned with {\em proof refinement}, pertinent to the Hilbert twenty-fourth problem (discovered by R. Thiele [Thiele2001]).
Both the preceding and the following (to me) are captured by the notion of seeking a missing proof of some type.
We have spent almost a decade, on and off, in the study of proof simplification (refinement) in various contexts, and we experienced approximately one year ago great satisfaction from Thiele's discovery and the following quote from Hilbert:
``The twenty-fourth problem in my Paris lecture was to be:  Criteria of simplicity, or proof of the greatest simplicity of certain proofs.''
 
A proof in hand can be simplified in many respects, and OTTER can provide substantial assistance in many of them.
Ceteris paribus, a reduction in the length of a proof corresponds to a simplification.
A reduction in the variable richness of a proof also contributes to its simplicity, where the variable richness equals the maximum number of distinct variables present among the deduced steps.
Similarly, a reduction in equational or formula complexity simplifies a proof, where the complexity measures in symbol count the longest deduced step.
In addition, simplicity is increased when so-called big lemmas are avoided and when various classes of term (such as double-negation terms) are avoided.
Still another aspect of proof simplification relates to proof size, the total number of symbols present in the deduced steps, a concept brought to our attention by D. Ulrich.
Each of these proof refinements has its analogue in the study of axiom systems.
For but one example, researchers sometimes pursue the discovery of an axiom system of smaller size than that in hand.
 
The majority of our research has focused on proof length, for which OTTER offers a number of methodologies.
Rather than detailing the various methodologies, we instead review the latest approach, in part because it illustrates well what can be accomplished with OTTER.
 
Imagine that the goal is to find a shorter proof of the conjunction of $A, B,$ and $C$ and that one has in hand a proof of said conjunction.
Next, let the proof of $C$ be the longest of the three subproofs respectively of $A$, $B$, and $C$.
Note the important fact that the goal of finding a shorter proof of the conjunction makes no demand on finding shorter proofs of any of the three members.
A strategy called {\em cramming} [Wos2001b] has proved quite powerful in this context, sometimes producing the desired shorter proof and at the same time relying on longer subproofs of one or more members of the conjunction.
 
The basic idea is to take the proof of $C$ and cram as many of its steps into the other two needed proofs as possible to thus require very, very few additional steps to reach the goal.
In the ideal case, the subproof of $C$ in hand is such that but two additional steps are required, one to deduce $A$, and one to deduce $B$.
In other words, the proof of $C$ offers all of the needed parents that permit an application of an inference rule in use to yield $A$ and another set to yield $B$.
If all goes as planned, the new proof of the conjunction is but two steps longer than the subproof of $C$.
One can test for this case by using a breadth-first search, adjoining the proof steps of $C$ to the initial set of support, placing in a hints list (by relying on R. Veroff's {\em hints strategy} [Veroff2001) $A$ and $B$, and assigning to max\_weight a very small value.
 
The ideal case has occurred in our research, producing an abridgment of a Meredith-Prior abridgment [Meredith1963,Wos2001b] for the proof for the {\L}ukasiewicz shortest single axiom for the implicational fragment of propositional calculus [{\L}ukasiewicz1970].
We have many other successes of using cramming in which the program was allowed more freedom but still keyed on the longest subproof of the members of the conjunction under study.
Other useful incarnations of cramming have been formulated and successfully applied.
Regarding trading short subproofs for longer ones, in one of those studies (of the <C,O> calculus [Meredith1953,Wos2001b], related to propositional calculus), cramming found a shorter proof (from a single axiom with the target a 2-axiom system) by trading a 10-step subproof of the second member for a 35-step proof of it.
 
If, instead of proof length, the simplification of concern is that of equational or formula complexity, OTTER offers the explicit means for attacking the problem, namely, the use of max\_weight.
When, say, the proof in hand has complexity $k$ and one seeks a proof of complexity $j$ with $j$ strictly less than $k$, one merely assigns $j$ as the value to max\_weight.
The program will then retain no new conclusions whose complexity exceeds $j$.
Similarly, if the proof in hand contains a deduced step that requires $k$ distinct variables and all other deduced steps require $k$ or fewer---its variable richness is $k$---one can easily search for a proof with strictly less richness by assigning a value less than $k$ to max\_distinct\_vars.
 
Still in the context of proof simplification, OTTER offers the means for seeking a proof that avoids the use of some thought-to-be-indispensable powerful lemma.
One merely instructs the program to reject if deduced the clause that corresponds to the unwanted lemma, either  through the use of demodulation or through the use of weighting.
Ordinarily, the absence of a powerful or deep lemma in a proof makes the proof simpler; indeed, one need not master the proof of the lemma.
Essentially the same approach can be applied if the refinement under consideration concerns some class of term that is to be avoided, for example, double-negation terms or terms containing as a proper subexpression $i(t,t)$ for any term $t$ and some function $i$.
In the context of an application outside of mathematics or logic, a circuit designer might wish to avoid nested {\em not} gates.
Such term avoidance, though often counterintuitive, can yield a simpler proof.
Of course, simplification in one property may be at the expense of simplification in another.
For example, blocking the use of a ``big'' lemma may result in a longer proof.
On the other hand, as occurred in our study of a dependency in infinite-valued sentential calculus, our methodologies applied by OTTER yielded a proof free of three lemmas used in the literature, 
free of double negation, and shorter than any proof of which we know, a proof of length 30 (applications of condensed detachment).
 
At this point, we turn from details about methodology to a brisk review of diverse successes.
We also include open questions to stimulate further research.
\section{Diverse Successes and Open Questions} %
 
This section offers a very small taste of what has been recently discovered with OTTER's assistance.
Bulleted items offer research topics.
 
Group theory has witnessed significant contributions by automated reasoning.
For the first such citation, consider groups of exponent 19, groups in which the nineteenth power of $x$ (for all elements $x$) is the identity $e$.
Such groups admit a single axiom, the following (in which the function $f$ denotes product).
\begin{verbatim}
(f(x,f(x,f(x,f(x,f(x,f(x,f(x,f(x,f(x,f(f(x,f(x,f(x,f(x,f(x,f(x,f(x,
f(x,f(x,f(f(x,y),z)))))))))),f(e,f(z,f(z,f(z,f(z,f(z,f(z,f(z,
f(z,f(z,f(z,f(z,f(z,f(z,f(z,f(z,f(z,f(z,z)))))))))))))))))))))))))))) = y).
\end{verbatim}
There does exist a shortest single axiom (proved by Kunen and Hart) obtained by dropping the occurrence of $e$ [Hart1995].
 
As for groups in general, single axioms have been respectively contributed by McCune [McCune1993] and by Kunen [Kunen1992], the first the shortest possible (proved by Kunen), and the second that with the least variable richness.
In the following, $f$ denotes product, and $g$ denotes inverse.
\begin{verbatim}
f(x,g(f(y,f(f(f(z,g(z)),g(f(u,y))),x)))) = u.
f(g(f(x,g(x))),f(f(g(x),y),g(f(g(f(x,z)),y)))) = z.
\end{verbatim}
\begin{itemize}
\item
Does there exist a single axiom whose length is that of the first cited, but whose variable richness is that of the second (3)?
\item
Does there exist a short single axiom for groups of exponent 6, those such that the sixth power of $x$ is the identity for all elements $x$?
Meredith has provided single axioms for groups of exponent 2 [Meredith1968], and Kunen has provided shortest single axioms for groups of exponent 4 [Kunen1995].
\end{itemize}
 
Lattice theory also has not escaped the consideration by OTTER in the context of single axioms.
Indeed, McCune has used this program to find the following 29-letter axiom, where v denotes join and  \verb|^| denotes meet.
\begin{verbatim}
(((y v x)^x) v (((z^ (x v x)) v (u^x))^v))^ (w v ((v6 v x)^ (x v v7)))=x.
\end{verbatim}
\begin{itemize}
\item
Does there exist a shorter single axiom for lattice theory?
\end{itemize}
 
Boolean algebra also relinquished some of its treasures to OTTER.
In particular, prompted by an e-mail in which S. Wolfram offered 25 candidate equations for being a single axiom, Veroff [Veroff2000] and McCune [McCune2001] conducted a study of that field in terms of the Sheffer stroke.
They proved two of the equations sufficient (including the following in which the function $f$ denotes the Sheffer stroke), proved that their mirror images are also sufficient, and (with colleagues) that seven are insufficient; the status of the remaining sixteen is still in doubt [McCune2001].
\begin{verbatim}
f(f(x,f(f(y,x),x)),f(y,f(z,x)))=y
\end{verbatim}
McCune in a separate study of Boolean algebra in terms of {\bf or} and {\bf not}, denoted by + and $\sim$, respectively, found ten single axioms, including the following.
\begin{verbatim}
~ (~ (~ (x + y) + z) + ~ (x + ~ (~ z + ~ (z + u)))) = z
\end{verbatim}

\begin{itemize}
\item
Does there exist a shorter single axiom in terms of disjunction and negation? (Colleagues have shown that no shorter single axiom in terms of the Sheffer stroke exists.)
\end{itemize}
 
Various fields of logic have also been successfully mined with OTTER.
The following new single axiom (in terms of the Sheffer stroke) for propositional logic was found by B. Fitelson.
\begin{verbatim}
P((D(D(x,D(y,z)),D(D(D(D(y,u),D(x,u)),D(u,y)),D(D(z,y),x))))).
\end{verbatim}
Fitelson then found the following and {\em first known} single axiom for $C4$ [Ernst2001].
\begin{verbatim}
P(i(i(x,i(i(y,i(z,z)),i(x,u))),i(i(u,v),i(w,i(x,v))))).
\end{verbatim}
\begin{itemize}
\item
Does there exist another single axiom for $C4$?
\end{itemize}

K. Harris then found a single axiom for the implicational fragment of infinite-valued sentential calculus, the following.
\begin{verbatim}
P(i(i(i(x,i(y,x)),i(i(i(i(i(i(i(i(i(z,u),i(i(v,z),i(v,u))),i(i(w,i(v6,w)),
v7)),v7),i(i(i(i(v8,v9),v9),i(i(v9,v8),v8)),v10)),v10),i(i(i(i(v11,v12),
i(v12,v11)),i(v12,v11)),v13)),v13),i(i(v14,i(v15,v14)),v16))),v16)).
\end{verbatim}

\begin{itemize}
\item
Does there exist a shorter single axiom for this area of logic?
\end{itemize}
 
Of a strikingly different nature are successes and questions focusing on proof refinement, pertinent to Hilbert's interest in proof simplification.
One interesting success concerns Kunen's shortest single axiom for groups (given earlier in this section).
Relying on a Knuth-Bendix approach and a corresponding input file supplied by Kunen, OTTER found a proof of length 98, a proof that includes more than fifty applications of demodulation.
Relying on various methodologies designed to yield (if possible) shorter proofs, and replacing Knuth-Bendix by a more standard use of paramodulation, OTTER eventually discovered a 43-step proof, a proof totally free of demodulation.

\begin{itemize}
\item
Does there exist a proof where the length is 42 or less?
\end{itemize}
 
Where diverse aspects of proof simplification are in focus, the Meredith single axiom for two-valued sentential calculus provided the wellspring for various successes.
His proof is (in effect) of length 41.
Our research has produced a 38-step proof.
\begin{itemize}
\item
Does there exist a shorter proof?
\end{itemize}

The Meredith proof has variable richness seven.
We have found a proof of richness five, which is the limiting case; that proof has length 68.
The Meredith proof relies on the use of double negation, containing seventeen steps of that type.
We have discovered a proof totally free of double negation, a proof of length 51.
\section{Summary and Invitation} %
 
The nature of research has changed.
Now one can choose to have the assistance of a powerful, general-purpose automated reasoning program.
OTTER, for example, offers a wide variety of strategies that enable the researcher to explore huge spaces of conclusions and traverse within that space areas that would be otherwise quite difficult, even counterintuitive, to explore.
One can use a reasoning program to find first proofs and settle conjectures.
Instead, one can enlist its assistance in proof simplification of diverse types (in the spirit of Hilbert's twenty-fourth problem).
 
An appealing aspect of the Hilbert-style, axiomatic proofs discovered by OTTER is the detail that is supplied.
Such proofs admit automated checking in most cases.
A researcher can learn from such proofs, and, as if some graduate student or colleague had provided the results of incomplete research, one can also learn from incomplete attempts by examining the conclusions that were drawn by the program in the attempt and placed in an output file.
Surprises occur.
For example, occasionally one finds that a thought-to-be-indispensable lemma is in fact not needed.
 
We invite suggestions in the realm of open questions where no proof exists or those focusing on some type of proof simplification.
The sampling we have presented in this article provides a clue concerning the nature of question that we have in mind.
Such questions or, for that matter, comment is welcome by surface mail or by e-mail, {\tt wos@mcs.anl.gov.}

\section*{References}

\vspace{.1in}
\noindent
[Epstein1995]  Epstein, R., {\em The Semantic Foundations of Logic: Propositional Logics,} 2nd ed., Oxford University Press, New York, 1995.

\vspace{.1in}
\noindent
[Ernst2001]  Ernst, Z., Fitelson, B., Harris, K., and Wos, L., ``Shortest Axiomatizations of Implicational S4 and S5'', Preprint ANL/MCS-P919-1201, December 2001.

\vspace{.1in}
\noindent
[Hart1995]  Hart, J., and Kunen, K., ``Single Axioms for Odd Exponent Groups'', {\em J. Automated Reasoning} {\bf 14}, no. 3 (1995) 383--412.

\vspace{.1in}
\noindent
[Kunen1992]  Kunen, K., ``Single Axioms for Groups'', {\em J. Automated Reasoning} {\bf 9}, no. 3 (1992) 291--308.

\vspace{.1in}
\noindent
[Kunen1995]  Kunen, K., ``The Shortest Single Axioms for Groups of Exponent 4'', {\em Computers and Mathematics with Applications} (special issue on automated reasoning) {\bf 29}, no. 2 (February 1995) 1--12.

\vspace{.1in}
\noindent
[{\L}ukasiewicz1970]  {\L}ukasiewicz, J., {\em Selected Works}, edited by L. Borokowski, North Holland, Amsterdam, 1970.

\vspace{.1in}
\noindent
[McCune1993]  McCune, W., ``Single Axioms for Groups and Abelian Groups with Various Operations'', {\em J. Automated Reasoning} {\bf 10}, no. 1 (1993) 1--13.

\vspace{.1in}
\noindent
[McCune1997]  McCune, W., ``Solution of the Robbins Problem'', {\em J. Automated Reasoning} {\bf 19}, no. 3 (1997) 263--276.

\vspace{.1in}
\noindent
[McCune2001]  McCune, W., Veroff, R., Fitelson, B., Harris, K., Feist, A., and Wos, L., ``Short Single Axioms for Boolean Algebra'', {\em J. Automated Reasoning}, accepted for publication.

\vspace{.1in}
\noindent
[Meredith1953] Meredith, C. A., ``Single Axioms for the Systems $\langle$C,N$\rangle$, $\langle$C,O$\rangle$, and $\langle$A,N$\rangle$ of the Two--Valued Propositional Calculus'', {\em J. Computing Systems} {\bf 1}, no. 3 (1953) 155--164.

\vspace{.1in}
\noindent
[Meredith1963]  Meredith, C. A., and Prior, A., ``Notes on the Axiomatics of the Propositional Calculus'', {\em Notre Dame J. Formal Logic} {\bf 4}, no. 3 (1963) 171--187.

\vspace{.1in}
\noindent
[Meredith1968]  Meredith, C. A., and Prior, A. N., ``Equational Logic'', {\em Notre Dame J. Formal Logic} {\bf 9} (1960) 212--226.

\vspace{.1in}
\noindent
[Thiele2001]  Thiele, R., and Wos, L., ``Hilbert's Twenty-Fourth Problem'', Preprint ANL/MCS-P899-0801, Mathematics and Computer Science Division, Argonne National Laboratory, Argonne, IL, 2001.

\vspace{.1in}
\noindent
[Wos1995]  Wos, L., ``The Resonance Strategy'', {\em Computers and Mathematics with Applications} {\bf 29}, no. 2 (February 1995) 133--178.

\vspace{.1in}
\noindent
[Veroff1996]  Veroff, R., ``Using Hints to Increase the Effectiveness of an Automated Reasoning Program: Case Studies'', {\em J. Automated Reasoning} {\bf 16}, no. 3 (1996) 223--239.

\vspace{.1in}
\noindent
[Veroff2000]  Veroff, R., ``Solving Open Questions and Other Challenge Problems Using Proof Sketches'', {\em J. Automated Reasoning} {\bf 27}, no. 2 (August 2001) 157--174.

\vspace{.1in}
\noindent
[Wos1999]  Wos, L., and Pieper, G. W., {\em A Fascinating Country in the World of Computing: Your Guide to Automated Reasoning,} World Scientific, Singapore, 1999.

\vspace{.1in}
\noindent
[Wos2001a]  Wos, L., ``Conquering the Meredith Single Axiom'',
{\em J. Automated Reasoning} {\bf 27}, no. 2 (August 2001) 175--199.

\vspace{.1in}
\noindent
[Wos2001b]  Wos, L., ``The Strategy of Cramming'', Preprint ANL/MCS-P898-0801, Mathematics and  Computer Science Division, Argonne National Laboratory, Argonne, Illinois, 2001.

\vspace{.1in}
\noindent
[Wos2002]  Wos, L., {\em Automated Reasoning and the Discovery of Missing and Elegant Proofs}, Rinton Press, to appear 2002.

\end{document}